\documentclass[letterpaper, 10 pt, journal, twoside]{IEEEtran}  

\IEEEoverridecommandlockouts                              

\usepackage{amsmath} 
\usepackage{amsfonts}
\usepackage{amssymb}  
\usepackage{dsfont}
\usepackage{graphicx}
\graphicspath{{figures/}}
\usepackage{subcaption}
\usepackage{array}
\usepackage{multirow}
\usepackage{hyperref}
\usepackage{cite}
\usepackage{comment}
\usepackage[ruled,linesnumbered]{algorithm2e}
\usepackage{amsthm}
\usepackage[utf8]{inputenc}
\usepackage[english]{babel}
\usepackage{siunitx}
\usepackage{tikz}
\usetikzlibrary{positioning}
\usetikzlibrary{arrows}
\usetikzlibrary{fit}
\usepackage{courier}  

\DeclareMathOperator{\reals}{\mathbb{R}}
\DeclareMathOperator*{\argmin}{\arg\!\min}

\usepackage{nccmath}  
\usepackage{mathtools}
\usepackage{xpatch}
\title{\Large \bf
A Kinematic Model for Trajectory Prediction \\
in General Highway Scenarios
}

\author{Cyrus~Anderson$^{1}$, Ram~Vasudevan$^{2}$, and Matthew~Johnson-Roberson$^{3}$
\thanks{This work was supported by a grant from Ford Motor Company via the Ford-UM Alliance under award N028603.}
\thanks{$^1$C. Anderson is with the Robotics Institute, University of Michigan, Ann Arbor, MI 48109 USA {\tt\footnotesize andersct@umich.edu}}%
\thanks{$^2$M. Johnson-Roberson is with the Department of Naval Architecture and Marine Engineering, University of Michigan, Ann Arbor, MI 48109 USA {\tt\footnotesize mattjr@umich.edu}}%
\thanks{$^3$R. Vasudevan is with the Department of Mechanical Engineering, University of Michigan, Ann Arbor, MI 48109 USA {\tt\footnotesize ramv@umich.edu}}%
}

\begin{document}
\maketitle

\begin{abstract}
Highway driving invariably combines high speeds with the need to interact closely with other drivers.
Prediction methods enable autonomous vehicles (AVs) to anticipate drivers' future trajectories and plan accordingly.
Kinematic methods for prediction have traditionally ignored the presence of other drivers, or made predictions only for a limited set of scenarios.
Data-driven approaches fill this gap by learning from large datasets to predict trajectories in general scenarios.
While they achieve high accuracy, they also lose the interpretability and tools for model validation enjoyed by kinematic methods.
This letter proposes a novel kinematic model to describe car-following and lane change behavior, and extends it to predict trajectories in general scenarios.
Experiments on highway datasets under varied sensing conditions demonstrate that the proposed method outperforms state-of-the-art methods.

\end{abstract}


\section{Introduction}

Highway driving places autonomous vehicles at high speeds and in close proximity to other drivers.
Conservative behaviors can completely eliminate the risk of collision in some scenarios~\cite{leung2020infusing}, but many demand close interaction with less conservative human counterparts.
We focus on prediction as an aid to assessing risk and navigating these scenarios.
Motion in the short-term is largely constrained by vehicle dynamics and simple kinematic models predict with fair accuracy in this setting~\cite{deo2018would}.
Further motion, however, receives greater influence from the driver's intent to maneuver or interact with other vehicles.
State-of-the-art methods train deep neural networks (DNNs) on large trajectory datasets to extract the subtle differences in motion to predict each maneuver.
While DNNs are capable of predicting trajectories in general scenarios, their complexity hinders interpreting both the learned parameters and the predictions.
This complexity is especially problematic for AVs that drive on behalf of human passengers.
Such a responsibility calls for models where the cause of erroneous predictions can be explained.
In offering transparency, this can help to make unknown risk known.
A complementary focus is to identify predictions that are not well-founded before their use, which can be accomplished by testing whether observed behaviors fall within modeling assumptions~\cite{gelman1996posterior,fan2001goodness}.
In the context of autonomous driving, testing can signal when the AV has encountered anomalous behavior, or that the current model is inadequate for the situation.
While such model validation has seen extensive development for model-based methods, there has been less for model-free methods.
The interpretable and model-based nature of kinematic methods is encouraging, but they fail to account for maneuvers and interactions that are not explicitly modeled. Their performance for long-term predictions in general scenarios has correspondingly not reached that of DNNs.
\begin{figure}[t]
  \centering
  \includegraphics[width=0.45\textwidth]{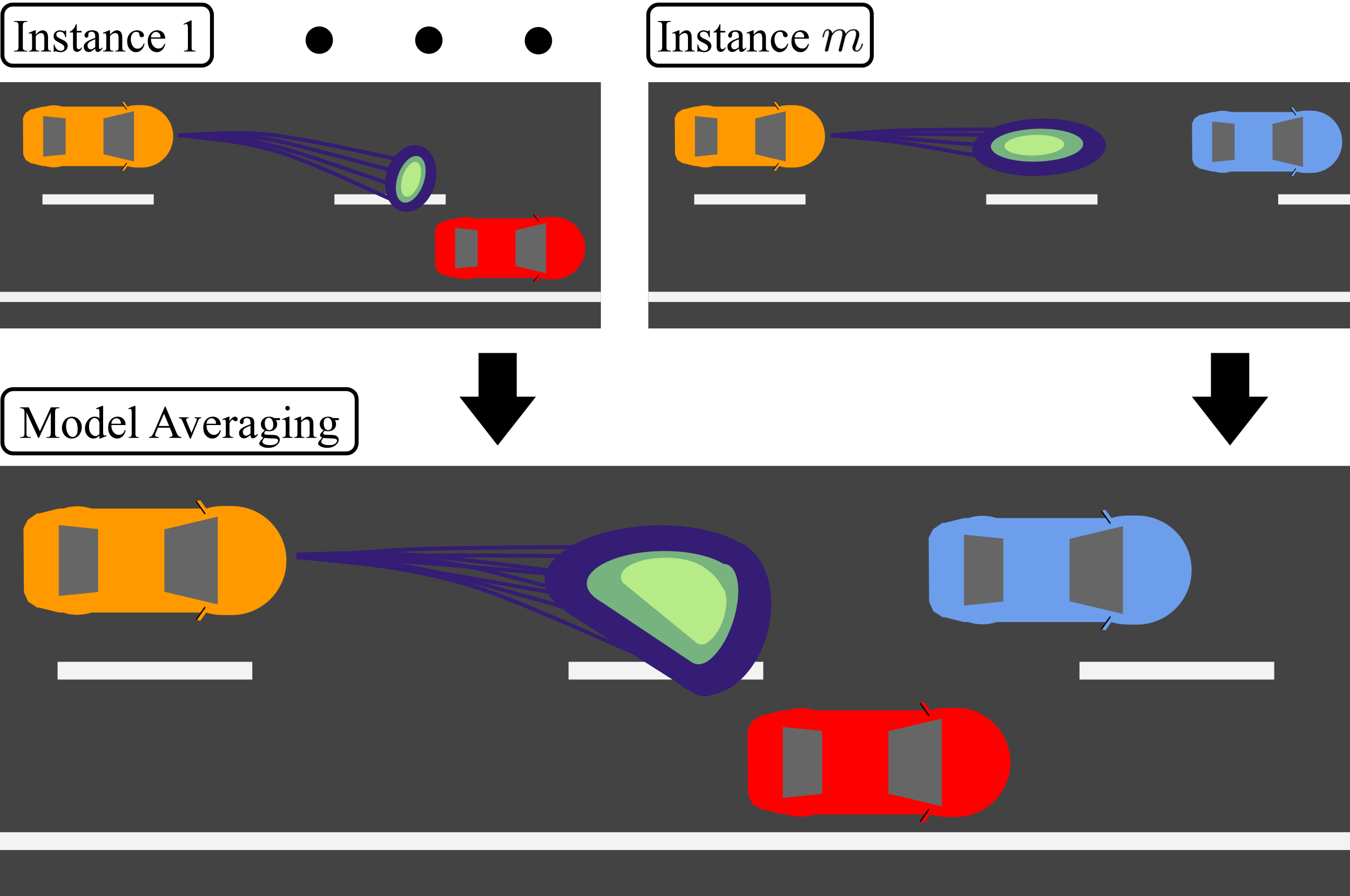}
  \caption{
Separate instances of the proposed kinematic model \textit{(top)} predict trajectories based on different interactions.
Combining these models enables predicting trajectories in general scenarios \textit{(bottom)}.
}
  \label{fig:overview}
\end{figure}
In this work we aim to build a method that is both interpretable and achieves high performance in general scenarios.
We approach this by first building a kinematic model that describes both lane change and car-following behavior.
The resulting model assumes that a target lane and a vehicle to follow have been specified.
To predict trajectories in general scenarios, we apply Bayesian model averaging over a suite of these models specified with different target lanes and vehicles.
The main contributions of this work are:
\begin{enumerate}
    \item a novel kinematic model for generating interaction-aware predictions in general highway driving scenarios;
    \item a tractable inference procedure;
	\item experiments on the highway traffic datasets NGSIM~\cite{ngsim} and highD\cite{krajewski2018highd} in varied sensing conditions.
\end{enumerate}
This letter is organized as follows.
Section~\ref{sec:related_works} describes related methods for trajectory prediction.
Section~\ref{sec:methods} introduces the proposed kinematic model and its extension for prediction in general highway scenarios.
We compare the proposed method to state-of-the-art approaches in Section~\ref{sec:experiments} and conclude in Section~\ref{sec:conclusion}.

\section{Related Work} \label{sec:related_works}

Methods that predict trajectories by specifying a kinematic model are described first.
The following section describes methods that rely on learning from large trajectory datasets to specify a model for prediction.

\subsection{Kinematic Methods}

The most basic kinematic models assume drivers move at a constant velocity, acceleration, or yaw rate~\cite{sorstedt2011new,schreier2016integrated,xie2017vehicle,deo2018would}.
These models achieve high accuracy over short time horizons due to their close approximation of true vehicle dynamics in common scenarios.
Over long time horizons, they often fail to predict drivers' intent to maneuver or interact with other vehicles.
One line of research has led to explicit models for lane change maneuvers based on single-integrator models~\cite{schreier2016integrated,xie2017vehicle} and optimizing over quintic polynomials~\cite{houenou2013vehicle}.
Another has instead focused on modeling interactions specifically in the case of car-following behavior.
Given a specified leader vehicle to follow, these methods predict the follower will maintain a desired distance and velocity~\cite{wei2013auto}, time gap~\cite{schreier2016integrated}, or a minimum distance subject to constraints on acceleration~\cite{treiber2000congested}.
While these works rely on manually chosen parameters, one approach to improve performance has been to estimate the parameters from observations made online~\cite{hoermann2018dynamic,sorstedt2011new,anderson2020lowlat}.
Online estimation adapts each model to each individual's driving, but generally requires solving a nonconvex problem. Proposed solutions include using particle filters~\cite{hoermann2018dynamic}, general purpose optimization routines~\cite{sorstedt2011new}, and convex relaxations based on semidefinite programs~\cite{anderson2020lowlat}. The computation involved in each approach hinders their inclusion in larger methods that model general highway driving.
In addition, car-following models say little about how to specify the leading vehicle. This is problematic in cases where the leading vehicle may merge out of the follower's lane, or when another vehicle begins to merge into the space between the two.
To resolve this, we instead treat the identity of the leader vehicle as unknown, and effectively estimate it online.

The above works largely focus on predicting specific maneuvers.
Prediction for general highway driving scenarios has been addressed by combining the results of maneuver-specific models.
Taking a linear combination is pursued within the framework of interactive multiple models~\cite{xie2017vehicle,deo2018would} and leads to a unimodal prediction when the predictions being combined are unimodal.
Bayesian model averaging~\cite{schreier2016integrated} instead weights each component prediction by its relative evidence and produces multi-modal predictions.
We adopt this latter approach to better predict multi-modal behavior.

\subsection{Data-Driven Methods}

In predicting arbitrary maneuvers, kinematic models are hampered by their need to explicitly model each maneuver and interaction considered.
Data-driven methods overcome this difficulty by leveraging large datasets to learn models for maneuvers and interactions from recorded trajectories.
Initial works aimed to learn maneuvers in a model-free framework via Gaussian mixture models (GMMs)~\cite{wiest2012probabilistic,deo2018would} or long short-term memory (LSTM) networks~\cite{xin2018nn}, and addressed interactions with manually specified cost functions~\cite{bahram2016combined,deo2018would}.
The need to manually specify cost functions, however, leads to the same difficulties faced by the kinematic models.
Subsequent works have aimed to learn models for maneuvers and interactions simultaneously, with many making use of the rich capacity for representation offered by deep neural networks (DNNs).
This has also spurred researchers to adapt DNNs to the task of trajectory prediction.
Deep neural networks in their base form are deterministic functions of their input, and some works have focused on accounting for the interactions between road users without modeling the uncertainty associated with future trajectory prediction~\cite{lee2017desire,gao2020vectornet}.
To adapt DNNs to describe uncertainty while still maintaining determinism, works have instead predicted the parameters corresponding to probability distributions.
Distributions predicted by DNNs include discrete distributions over a finite number of trajectories~\cite{liang2020learning} and normal distributions for each predicted timestep~\cite{alahi2016social,chandra2019traphic}.
Deo \textit{et al.}~\cite{deo2018nn} achieve both a multi-modal and continuous description of uncertainty by predicting a normal distribution for each type of maneuver from a given class.
This makes predictions following a Gaussian mixture model, but relies on maneuvers to be labeled.
Other works have aimed to predict GMMs without the aid of labeled data~\cite{chai2020multipath,mercat2020multi}.
Training a mixture model without the structure provided by maneuver labels, however, may collapse separated modes into a single mode during the training process~\cite{jain2020discrete}.
Chai \textit{et al.}~\cite{chai2020multipath} propose a two-step procedure to train a mixture model and DNN separately but are unable to benefit from end-to-end training.
Occupancy grids present an alternative approach to predict multi-modal distributions~\cite{park2018sequence,jain2020discrete} but entail a trade-off between the discretization error of a coarse grid and the increased computation imposed by a fine grid.
While the above methods predict the parameters that exactly specify a given possibly discrete distribution, a number of methods instead rely on learning a latent distribution from which predicted trajectories are drawn as samples~\cite{choi2020shared,li2020wagdat,salzmann2020trajectron,gupta2018socialgan,zhao2019multi,huang2020diversitygan}.
Several of these works use the variational autoencoder framework to learn a latent distribution over interactions, and model each pairwise interaction between road users~\cite{choi2020shared,li2020wagdat,salzmann2020trajectron}.
To reduce the computation involved in the pairwise models, one solution has been to consider interactions only between road users within a fixed distance~\cite{salzmann2020trajectron}.
Social GAN~\cite{gupta2018socialgan} instead proposes a novel pooling module to examine interactions without the need to form all pairs, thereby removing the issues of computational complexity and manual choice of distance threshold.
The price of this pooling module is that the spatial relations between road users are lost when using the single pooled result to make predictions.
Zhao \textit{et al.}~\cite{zhao2019multi} achieve state-of-the-art performance by preserving spatial relations within a tensor that models an inertial frame.
Since sampling each model's latent distribution can be costly, DiversityGAN~\cite{huang2020diversitygan} introduces a low-dimensional space to more efficiently sample rare events such as lane changes. Constructing this space, however, depends on labeled data.

\section{Probabilistic Trajectory Predictions} \label{sec:methods}

The proposed kinematic model consists of separate longitudinal and lateral components.
Section~\ref{ssec:problem_statement} states the trajectory problem for which it is built, and provides an overview of the model.
The following sections describe the longitudinal component (Section~\ref{ssec:lon_model}) and the lateral component (Section~\ref{ssec:lat_model}).
The extension to predicting trajectories in general highway scenarios is described in Section~\ref{ssec:general_scenarios}, and a tractable inference procedure is detailed in Section~\ref{ssec:inference}.

\subsection{Problem Statement}\label{ssec:problem_statement}

We observe the target vehicle's position during a window lasting $n$ timesteps and aim to predict its position until a final timestep $T$.
Let the subset of observed timesteps be given by $S \subseteq \{1,...,n\}$, where we assume $1 \!\in\! S$ without loss of generality.
The position at timestep $t$ in the ground plane is denoted $p(t) \!\in\! \reals^2$.
As in other works~\cite{sorstedt2011new,schreier2016integrated,xie2017vehicle} we assume the position is given in terms of longitudinal $p_1(t)$ and lateral $p_2(t)$ coordinates along the road with $p(t) \!=\! (p_1(t), p_2(t))^\intercal$.
In the general case, positions can be transformed to this coordinate system using low-degree polynomial models of the road~\cite{kim2015curvilinear}.
We also assume we observe the positions of $n_v$ other vehicles over the same observation window.
The $j$th vehicle's position at timestep $t$ is denoted $p^j(t)$.
For convenience we denote the collection of other vehicles' observations by $V \!=\! \{ \{p^j(t)\}_{t=1}^n \,|\, j=1,\dots,n_v \}$, the target vehicle's observed positions by $p_S \!=\! \{p(t)\}_{t\in S}$, and future positions by $p_T \!=\! \{p(t)\}_{t=n+1}^T$.
We now write the prediction task as sampling the target vehicle's future trajectories based on the observed data:
\begin{align} \label{eqn:problem_statement}
    p_T \sim \Pr(p_T \,|\, p_S, V).
\end{align}
The kinematic model used to make predictions relies on driving that is away from the limits of handling, such that vehicle dynamics for lateral and longitudinal motion are approximately decoupled.
Since this applies to most highway driving, we model each component as a separate double-integrator.
Given $\Delta t$ as the size of each timestep and the component in $i\!=\!1,2$, motion is given by: 
\begin{equation} \label{eqn:dynamics}
    x_i(t+1) = Ax_i(t) + B(u_i(t) + \varepsilon_i(t)) ,
\end{equation}
where 
\begin{equation}
    A = \begin{pmatrix} 1&\Delta t\\0&1 \end{pmatrix}, 
    B = \begin{pmatrix} 0\\1 \end{pmatrix},
\end{equation}
with $u_i(t)$ as a modeled control input and $\varepsilon_i(t)$ is the Gaussian white noise with variance $\sigma_i^2$ that is specific to each component.
This model may produce negative velocities in stop-and-go traffic, so for the propagation of state over the prediction horizon $t\!=\!n,\dots T$ we set longitudinal velocities that would become negative to zero.
In addition, the surrounding vehicles' positions are assumed given only over the observation window. For prediction, their states are propagated according to~\eqref{eqn:dynamics} with zero control input.

\subsection{Longitudinal Motion Model}\label{ssec:lon_model}

The vehicle's longitudinal motion is described by a car-following model, based on maintaining a desired gap $g*$ and desired velocity $v*$ similar to previous work~\cite{wei2013auto,anderson2020lowlat}.
We treat these quantities as unknown, and aim to infer them from observed interactions with the leading vehicle.
Since the desired values likely vary in time, we approximate them as constant and assume the vehicle is closest to achieving these at the end of the observation window.
This results in the priors:
\begin{gather}
    g^* \sim N(p^i_1(n) - p_1(n), \sigma_g^2), \\
    v^* \sim N(v_1(n), \sigma_v^2),
\end{gather}
where the $i$th vehicle is given as the leading vehicle.
The longitudinal control is assumed to be calculated over a fixed horizon $k_f$.
Let $u \!=\! (u_0, \dots, u_{k_c-1}) \!\in\! \reals^{k_f}$ and let $x_{1,0}$ denote the current longitudinal state of the target vehicle.
In addition, let $x^i_{1,0} \!=\! (p^i_{1,0}, v^i_{1,0})^\intercal$ denote the leading vehicle's current state.
The longitudinal control $u_1(t)$ is given by the control law $u_1(t) = u_0$ where $u$ solves:
\begin{equation} \label{eqn:opt_lon}
\begin{aligned}
    \underset{\substack{
        u \in \reals^{k_f} 
    }}{\mathrm{minimize}}& ~~ \lVert u \rVert_2^2 & \\
    \mathrm{s.t.} ~~ & x_1(k_c) = (p^i_{1,0} + k_c \Delta t v^i_{1,0}- g^*, v^*)^\intercal & \\
    & x_1(0) = x_{1,0}  & \\
    \forall\, i=0,\dots,k_f\!-\!1 ~~ & x_1(i\!+\!1) = Ax_1(i) + Bu_i & \\[-8pt]
\end{aligned}
\end{equation}
wherein the driver assumes the leading vehicle moves at a constant speed.
In the case of no lead vehicle, $u_2(t)$ is given by:
\begin{align}
    u_1(t) = \frac{v^* - v_1(t)}{k_f}.
\end{align}
The fact that the control is depends linearly on the state and the variables $g^*,v^*$ is later used during inference.

\subsection{Lateral Motion Model}\label{ssec:lat_model}
Real drivers decide at discrete times to change lanes.
Rather than explicitly modeling a switching process we will use an approximate form that assumes there exists exactly one lane change that is partially observed. We make the same assumption as in other works~\cite{schreier2016integrated,xie2017vehicle} that after the lane change ends, the driver continues within the same lane.
Since the duration of a typical lane change is between four and ten seconds~\cite{thiemann2008estimating}, and a standard observation window is only three seconds, the majority of merges will be only partially observed.
We start by defining the set $\mathcal{K}$ to contain the possible durations of timesteps that remain in the lane change maneuver.
The unknown quantities in the model are the target vehicle's actual duration $k$ remaining in the lane change at $t=1$, and the desired lateral position $p_m$.
As an uninformative prior we assume the duration is uniformly distributed: 
\begin{align}
    k \sim U(\mathcal{K}).
\end{align}
The desired lane is assumed to be given with lateral center $\mu_p \!\in\! \reals$, with which we assume the desired lateral position $p_m$ is normally distributed about the center:
\begin{align}
    p_m \sim N(\mu_p, \sigma_p^2).
\end{align}
Since the target vehicle switches to continuing within the lane at some time, we now define the horizon used by the controller at each timestep $t$ as:
\begin{align}
    k_t =
    \begin{cases}
    k - t, & \text{if } k - t > 2\\
    k_s,              & \text{otherwise}
    \end{cases}
\end{align}
where $k_s$ is the fixed horizon used for continuing within the same lane.
We note that lane keeping behavior corresponds to choosing $p_m$ within the current lane.
To define the control input, let $u \!=\! (u_0, \dots, u_{k_t-1}) \!\in\! \reals^{k_t}$ and let $x_{2,0}$ denote the current lateral state. The lateral control $u_2(t)$ is given by the time-varying control law $u_12t) \!=\! u_0$ where $u$ solves:

\begin{equation} \label{eqn:opt_lat}
\begin{aligned}
    \underset{\substack{
        u \in \reals^{k_t} 
    }}{\mathrm{minimize}}& ~~ \lVert u \rVert_2^2 & \\
    \mathrm{s.t.} ~~ & x_2(k_t) = (p_m, 0)^\intercal & \\
    & x_2(0) = x_{2,0}  & \\
    \forall\,i=0,\dots,k_t-1 ~~ & x_2(i+1) = Ax_2(i) + Bu_i & \\
\end{aligned}
\end{equation}
Though $k$ is unknown, for inference we make use of the fact that the control is a linear function of state and $p_m$.

\begin{figure}[t]
  \centering
  \includegraphics[width=0.45\textwidth]{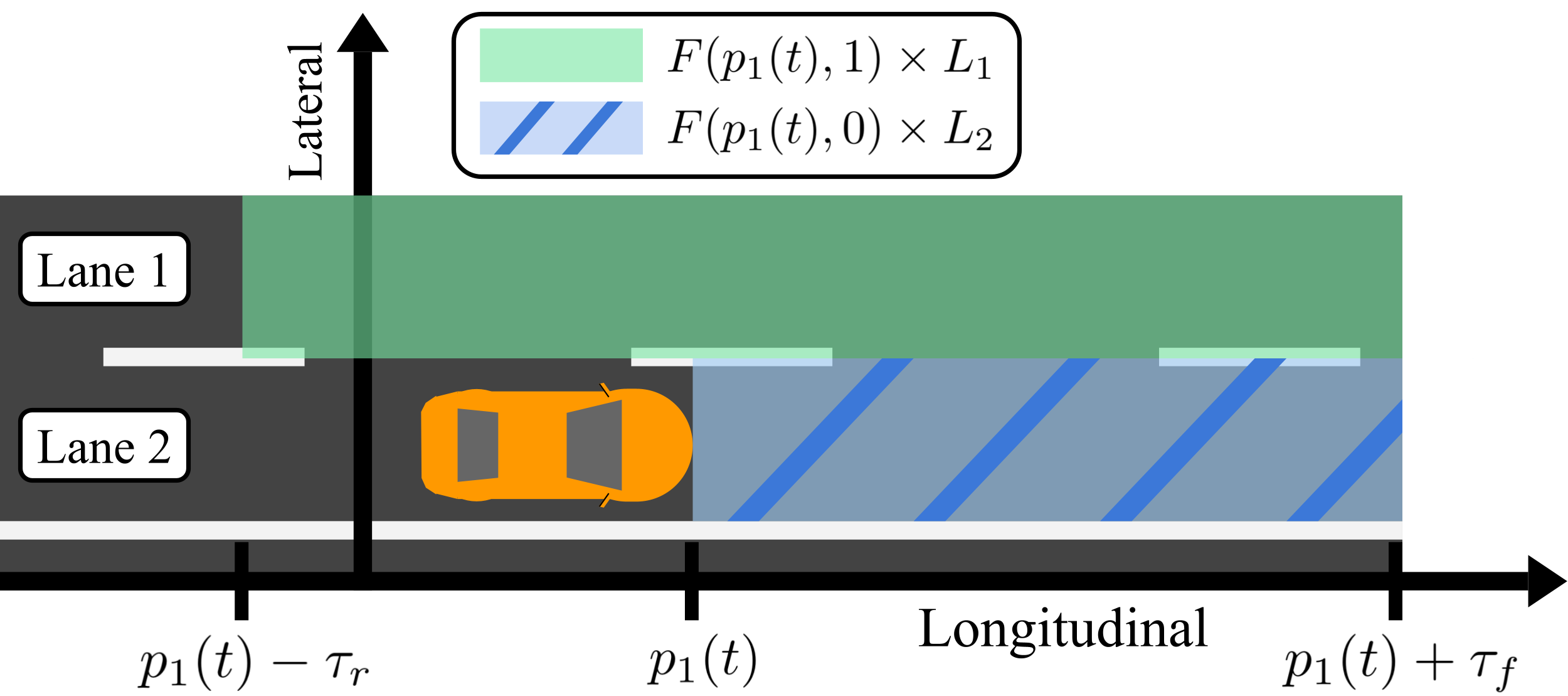}
  \caption{Target vehicle's field of view across two lanes (shaded) at timestep $t$.
  The view is specified in the lateral direction by the lane extents $L_1$ and $L_2$, and in the longitudinal direction by the current position $p_1(t)$, forward distance $\tau_f$, and the rear distance $\tau_r$.
}
  \label{fig:set_diagram}
\end{figure}

\subsection{General Highway Predictions}\label{ssec:general_scenarios}

Combining the lateral and longitudinal models from the previous sections, we must specify a target lane and a leading vehicle to predict future trajectories. Trajectories may then be sampled from the model written as:
\begin{align} \label{eqn:single_model}
    p_T \sim \Pr(p_T \,|\, p_S, V, i, j),
\end{align}
where the $i$th lane and the $j$th vehicle have been specified, and for notational convenience, we allow $j=\emptyset$ to denote the case of no lead vehicle for the longitudinal motion model.
Although model~\eqref{eqn:single_model} may be used to predict trajectories, it requires information that we are unlikely to know: the driver's desired lane and the vehicle to which they adjust their driving.
In this section we remove such a need by first identifying a set $C$ containing all relevant $(i, j)$ lane and lead vehicle pairs to consider.
Each pair is then used to specify a single model, and we apply Bayesian model averaging to combine them as:
\begin{align} \label{eqn:bma}
\begin{split}
    \Pr(p_T \,|\, p_S, V&) = \\
    \sum_{(i, j) \in C}  & \Pr(p_T \,|\, p_S, V, i, j)\Pr(i, j \,|\, p_S, V)
\end{split}
\end{align}
with predictions given by~\eqref{eqn:problem_statement}.
This decomposition shows that the averaged model takes the trajectories predicted by each component model and weights them by the component's evidence.
To build the set $C$ we first introduce the sets that describe the target vehicle's field of view.
Let the interval for $i$th lane's lateral values $[p_{i,l},p_{i,u}]$ be given by $L_i$.
For the longitudinal extent seen by the target vehicle, we introduce a forward distance $\tau_f$ and a rear distance $\tau_r$.
Here we assume that for the purposes of car-following, the target vehicle ignores those directly behind it.
Then letting $p$ denote the target vehicle's longitude, we define the extent as:
\begin{align}
    F(p,q) = [p-q\tau_r, p+\tau_f],
\end{align}
where $q$ is a binary value used to exclude the rear portion of the target vehicle's lane.
Figure~\ref{fig:set_diagram} shows the target vehicle's view modeled over two lanes.
Assume the first observation $p(1)$ corresponds to lane $l$.
Given the $i$th lane we can now write its set of paired lead vehicles by:
\begin{align}
\begin{split}
    G(i) = \{ j \in \{1,\hdots &,m\} \, | \,
    \exists t_1,t_2 \in S,~ p_2^j(t_2) \in L_i, \\
    & p_1^j(t_1) \in F(p_1(t_1), \mathds{1}\{l \neq i\})
    \}
\end{split}
\end{align}
which ensures each lead vehicle has been observed within each extent of the target vehicle's field of view at least once.
Denoting the set of lanes adjacent to lane $l$ by $\mathrm{adj}(l)$, we consider only lanes $i \in \{l\} \cup \mathrm{adj}(l)$.
The set of lane and lead vehicle pairs is now given by:
\begin{align}
    C = \{i,j \,| \, j \in G(i)\} \cup \{i,\emptyset \,| \, G(i) = \emptyset \},
\end{align}
where the lanes with no lead vehicles are paired in the second set. The next section describes a tractable method to predict trajectories using the combined model.

\subsection{Inference}\label{ssec:inference}

To sample trajectory predictions, we start by making use of the structure within the component longitudinal and lateral models given in Section~\ref{ssec:lon_model} and Section~\ref{ssec:lat_model} respectively.
The longitudinal model depends on the desired gap $g^*$ and velocity $v*$, and the lateral model depends on the target lateral position $p_m$ and merge duration $k$.
The key to efficient inference is that fixing $k$ makes the model's control inputs linear functions of the remaining unknowns.
To capitalize on this observation, we integrate over both $k \in \mathcal{K}$ and $(i,j) \in C$, and estimate the remaining parameters via Kalman filtering.
This process is explained next.
We first collect the random variables needed for prediction as $\theta = (x(n), g*, v*, p_m) \in \reals^7$.
To obtain a representation more amenable to inference, we rewrite~\eqref{eqn:bma} as:
\begin{align} \label{eqn:infer1}
    \Pr(p_T \,|\, p_S, V) = 
    \sum_{i, j, k} \int \Pr(p_T, \theta, i,j,k \,|\, p_S, V) d\theta .
\end{align}
The summand can be decomposed using the chain rule as:
\begin{align} \label{eqn:infer2}
\begin{split}
    \Pr(p_T, &\, \theta, i,j,k \,|\, p_S,V) = \\
    & \Pr(p_T \,|\, \theta, i,j,k,p_S,V)\Pr(\theta \,|\, i,j,k, p_S,V) \\
    & \Pr(i,j,k \,|\, p_S,V).
\end{split}
\end{align}
The first term in the chain represents the prediction of future trajectories by propagating the current state estimate $x(n)$.
Since the current state is included in $\theta$, conditional independence implies that we may remove $p_S$.
The second term represents the uncertainty in estimating the current state $x_n$ along with the other unknown parameters, and is exactly the posterior distribution estimated by the Kalman filter.
The final term can be rewritten using Bayes' rule:
\begin{align}
    \Pr(i,j,k \,|\, p_S,V) &= \frac{\Pr(p_S \,|\, i,j,k,V)\Pr(i,j,k \,|\, V)}{\Pr(p_S \,|\,V)} \\
    &\propto \Pr(p_S \,|\, i,j,k,V)\Pr(i,j,k \,|\, V) \\
    &= \Pr(p_S \,|\, i,j,k,V), \label{eqn:infer3}
\end{align}
where the final equality follows from assuming a uniform distribution over $(i, j, k) \!\in\! C \times \mathcal{K}$, independent of $V$.
The resulting term is the marginal probability of the observations $p_S$ under the model specified by $i,j,k$. 
We are now in a position to simplify~\eqref{eqn:infer2} and apply~\eqref{eqn:infer3} as:
\begin{align} \label{eqn:infer4}
\begin{split}
    \Pr(p_T, &\, \theta, i,j,k \,|\, p_S,V) \propto \\
    & \Pr(p_T \,|\, \theta, i,j,k,V)\Pr(\theta \,|\, i,j,k, p_S,V) \\
    & \Pr(p_S \,|\, i,j,k,V).
\end{split}
\end{align}
This decomposition suggests the following procedure sample trajectories from~\eqref{eqn:infer1}.
For each model component specified by $(i,j,k) \!\in\! C \!\times\! \mathcal{K}$, we use Kalman filtering to compute the last two terms in~\eqref{eqn:infer4}.
A value of $\theta$ can be sampled from its posterior distribution, and propagating the sampled state $x(n)$ yields a sample of $p_T$.
These predictions are then weighted by the marginal probability, and normalized by the sum of marginals across all components.
This inference procedure is summarized in Algorithm~\ref{algo:sampler}.
We note that if the constraint on longitudinal velocities to be nonnegative was removed, the propagation would also be possible within a Kalman filter for each model component.
Performing inference with standard filtering recursions allows the predictions for most scenarios to be made with a small number of filtering steps.
Scenarios for which $C$ remains constant from the previous timestep will require only $|C \times \mathcal{K}|$ filtering steps, one for each component model, in addition to the steps used to propagate the trajectories into the future.
Furthermore, each component model may be updated in parallel.

\begin{algorithm}[t!]
\SetAlgoLined
\KwIn{$p_S$, $V$, $C$, $\mathcal{K}$, $n$, $T$}  
\KwOut{Sampled trajectories $\{w^{(l)}, p_T^{(l)}\}$}
Let $m = |C \times \mathcal{K}|$ \\
\For{$l = 1,\dots,m$}{
  Take $l$th $(i,j,k)$ from $C \times \mathcal{K}$  \\
  Compute the posterior and marginal probabilities in~\eqref{eqn:infer4} using a Kalman filter with model~\eqref{eqn:dynamics} specified with $(i,j,k)$. \\
  Sample $\theta^{(l)} \sim \Pr(\theta \,|\, i,j,k, p_S,V)$ \\
  Propagate $\theta^{(l)}$ to obtain $p_T^{(l)}$ \\
  $\hat{w}^{(l)} \leftarrow \Pr(p_S \,|\, i,j,k,V)$ \\
}
$\forall\,l=1,\dots,m ~~ w^{(l)} \leftarrow \hat{w}^{(l)}/\sum_{i=1}^m \hat{w}^{(i)}$ \\
\caption{Inference for Trajectory Prediction}
\label{algo:sampler}
\end{algorithm}

\section{Experiments} \label{sec:experiments}
We evaluate the proposed method's ability to predict trajectories with two highway traffic datasets.
The first is the NGSIM~\cite{ngsim} dataset, which contains over 9,000 unique vehicles recorded at \SI{10}{\Hz} during dense and occasionally stop-and-go traffic at two highways in California.
The second dataset is highD\cite{krajewski2018highd}, which puts greater focus on general driving conditions.
It contains over 110,000 unique vehicles and is recorded at \SI{25}{\Hz} across six German highways near Cologne.
In each experiment we aim to predict five seconds into future based on a three second window of observations, as in other works~\cite{deo2018nn,chandra2019traphic,zhao2019multi}.

\subsection{Model Specifications}

To match NGSIM we use a timestep of \SI{0.1}{\s} for the proposed model.
For lane change duration $\mathcal{K}$ we use a grid of values between zero and 12 seconds with a spacing of \SI{0.5}{\s}.
Each driver is also assumed to plan their control inputs for non-merge situations over a ten second horizon, with $k_s \!=\! k_c \!=\! 100$.
Since lane widths on highways are commonly between \SI{3.5}{\meter} and \SI{4.5}{\meter}, the prior uncertainty $\sigma_p$ for desired offset to the lane center is set to \SI{1.5}{\meter}.
The uncertainties $\sigma_g, \sigma_v$ for desired gap and velocity are both set to \SI{2}{\meter} to provide some prior information. Previous work has shown that providing a small amount of information in the prior distribution can aid in predictions when estimating car-following models online~\cite{anderson2020lowlat}.
For the error introduced into control inputs in~\eqref{eqn:dynamics}, we set the lateral error $\sigma_2$ to \SI{0.05}{\meter} for all experiments.
Since NGSIM consists primarily of dense traffic, we set longitudinal error $\sigma_1$ to \SI{0.2}{\meter} for NGSIM and otherwise equal to $\sigma_2$.
The view distances $\tau_f$ and $\tau_r$ are set to \SI{50}{\meter} and \SI{10}{\meter} respectively.

\begin{table*}[t!]
\centering
\caption{Performance on NGSIM and highD datasets shown as average\,/\,final time error in meters (best in \textbf{bold} and second best \underline{underlined}).}
\begin{tabular}{ |c|c|c|c|c|c|c|c| }
\hline
    \multicolumn{8}{|c|}{Bird's Eye View Predictions}\\
\hline
  Dataset & Metric  & CV  & Social LSTM & Social GAN  & MATF & Proposed-NI & Proposed \\
\hline
     & QDE\,(0.2) & 1.99/3.86 & \textbf{1.70}/\underline{3.23} & 1.82/\textbf{3.20} & 2.40/4.59 & 2.01/3.93 & \underline{1.75}/3.42 \\
   NGSIM & ADE  & 3.56/6.90 & 4.10/7.86 & \underline{2.85}/\underline{5.39} & \textbf{2.50}/\textbf{4.76} & 3.69/7.36 & 3.14/6.18 \\
         & RMSE & 4.47/8.64 & 5.17/9.86 & \underline{3.76}/\underline{7.08} & \textbf{3.41}/\textbf{6.48} & 4.67/9.25 & 4.08/7.97 \\
\hline
     & QDE\,(0.2) & 1.20/2.37 & 1.31/2.61 & 2.11/3.89  & 1.94/3.75 & \underline{1.05}/\underline{2.17} & \textbf{0.99}/\textbf{2.08}  \\
   highD & ADE  & 2.56/5.04 & 2.42/4.87 & 3.03/5.89  & 2.11/4.02 & \underline{1.79}/\underline{3.74} & \textbf{1.51}/\textbf{3.16}  \\
         & RMSE & 3.14/6.22 & 3.64/7.09 & 6.91/12.57 & 4.74/8.76 & \underline{2.24}/\underline{4.75} & \textbf{1.92}/\textbf{4.04}  \\
\hline
\hline
    \multicolumn{8}{|c|}{Driver View Predictions}\\
\hline
  Dataset & Metric  & CV  & Social LSTM & Social GAN  & MATF & Proposed-NI & Proposed \\
\hline
     & QDE\,(0.2) & 2.12/4.02 & \textbf{1.83}/\textbf{3.41}  & 2.47/4.54 & 3.27/6.42 & 2.21/4.21 & \underline{1.98}/\underline{3.76} \\
   NGSIM & ADE  & 3.80/7.20 & 4.24/8.06  & \underline{3.50}/\textbf{6.64} & \textbf{3.39}/\textbf{6.64} & 4.01/7.79 & 3.53/\underline{6.78} \\
         & RMSE & 4.86/9.08 & 5.39/10.15 & \underline{4.61}/\underline{8.60} & \textbf{4.52}/\textbf{8.71} & 5.18/9.85 & 4.67/8.78 \\
\hline
     & QDE\,(0.2) & \textbf{1.27}/\textbf{2.51} & 1.55/3.07 & 2.91/5.26  & 2.50/4.84  & 1.50/3.13 & \underline{1.30}/\underline{2.70}  \\
   highD & ADE    & 2.68/5.25 & 2.73/5.42 & 3.90/7.39  & 2.68/5.13  & \underline{2.36}/\underline{4.96} & \textbf{1.88}/\textbf{3.90}  \\
         & RMSE   & 3.33/6.52 & 4.09/7.87 & 8.35/15.02 & 5.57/10.33 & \underline{3.08}/\underline{6.47} & \textbf{2.44}/\textbf{5.06}  \\
\hline
\end{tabular}
\label{tab:traditional_test}
\end{table*}

\subsection{Baselines}

We compare to prediction methods including DNNs that achieve state-of-the-art performance on the NGSIM dataset:
\begin{itemize}
    \item \textbf{Constant Velocity (CV)}\textbf{:} Vehicle motion is modeled by constant velocity.
    \item \textbf{Social LSTM (SLSTM)}\cite{alahi2016social}\textbf{:} An LSTM framework models the influence of nearby vehicles using a grid to define a social pooling module. 
    \item \textbf{Social GAN (SGAN)}\cite{gupta2018socialgan}\textbf{:} A GAN architecture that uses a pooling operator to incorporate all road users' interactions at once to provide context for predictions.
    \item \textbf{Multi-Agent Tensor Fusion (MATF)}\cite{zhao2019multi}\textbf{:} Drivers' spatial interactions are treated within a tensor that models single global frame to preserve context information.
    \item \textbf{No Interaction (Proposed-NI)}\textbf{:} A variant of the proposed method where interactions are ignored. This treats the set of observations of surrounding vehicles as empty.
\end{itemize}
The open source implementation for each DNN is trained on a separate set of data than that used for evaluation.
For NGSIM, each method is trained on data recorded at I-80 then evaluated on data at US-101, and vice-versa.
The highD data are separated into one split containing highways labeled one to three and another split containing the remaining highways, labeled four to six.
The two splits are then used in the same fashion as the two highway datasets in NGSIM.
Since highD is recorded at \SI{25}{\Hz}, it is resampled to match NGSIM at \SI{10}{\Hz}.
The DNNs operate on data at a lower frequency, so the input provided during evaluation and training is downsampled to \SI{5}{\Hz} for MATF and \SI{2}{\Hz} for Social LSTM and Social GAN.

\subsection{Evaluation Metrics}
We evaluate how well each method predicts the future with several metrics for probabilistic methods.
Let $p_{i,t}$ denote the $i$th vehicle's true position at timestep $t$, with the random variable corresponding to its prediction as $\hat{p}_{i,t}$.
For each DNN we sample 100 trajectory predictions to fully evaluate the posterior predictive distribution of $\hat{p}_{i,t}$.
Let $N$ be the total number of evaluated vehicles.
We calculate the following metrics at each second in the five second prediction window, and compare their time average along with their final value:
\begin{itemize}
    \item \textit{Root Mean Squared Error (RMSE):} The square root of expected squared distance between the true position and prediction, used in~\cite{deo2018nn,chandra2019traphic,zhao2019multi}. RMSE at timestep $t$ is given by: 
    \begin{align}
        RMSE(t) = \sqrt{\frac{1}{N}\sum_{i=1}^{N} \mathbb{E}[\lVert p_{i,t} - \hat{p}_{i,t}\rVert_2^2]}
    \end{align}

    \item \textit{Average Distance Error (ADE):} The expected distance between the true position and prediction, used in~\cite{alahi2016social,salzmann2020trajectron,gupta2018socialgan, zhao2019multi}. ADE at timestep $t$ is: 
    \begin{align}
        ADE(t) = \frac{1}{N}\sum_{i=1}^{N} \mathbb{E}[\lVert p_{i,t} - \hat{p}_{i,t}\rVert_2]
    \end{align}
    \item \textit{Quantile Distance Error (QDE):} The smallest distance traveled from the true position needed to reach a given fraction of the predictions. The value at timestep $t$ with fraction $q$ is: 
    \begin{align}
        QDE(q,t) = \frac{1}{N}\sum_{i=1}^{N} d_i,
    \end{align}
    where the distance $d_i$ is given by:
    \begin{align}
    \begin{split}
        d_i = & \argmin d  \\ & ~~ \mathrm{s.t.}~ \Pr(\lVert p_{i,t} - \hat{p}_{i,t}\rVert_2 \leq d) \geq q 
    \end{split}
    \end{align}
\end{itemize}
The quantile metric is equivalent to the minimum-of-K metrics~\cite{gao2020vectornet,alahi2016social,chai2020multipath,salzmann2020trajectron,gupta2018socialgan,zhao2019multi,huang2020diversitygan} when predictions are weighted by the same probability.
Measuring the quantile based metric favors predictions that place significant probability mass near the true position, without penalizing additional predictions that may be distant.
From the perspective of autonomous vehicles, this is the most relevant metric when we prioritize conservative driving.
In contrast, the expectation based metrics place more emphasis on predictions that cluster near the true position.
The dependence of RMSE on the squared error also makes it more sensitive to distant predictions.
This focus on overall closeness may be more desirable when the aim is to avoid distant predictions that could induce sudden and unwarranted emergency maneuvers.

\subsection{Bird's-Eye View Predictions}
\begin{figure*}[h!]
\includegraphics[width=0.95\textwidth]{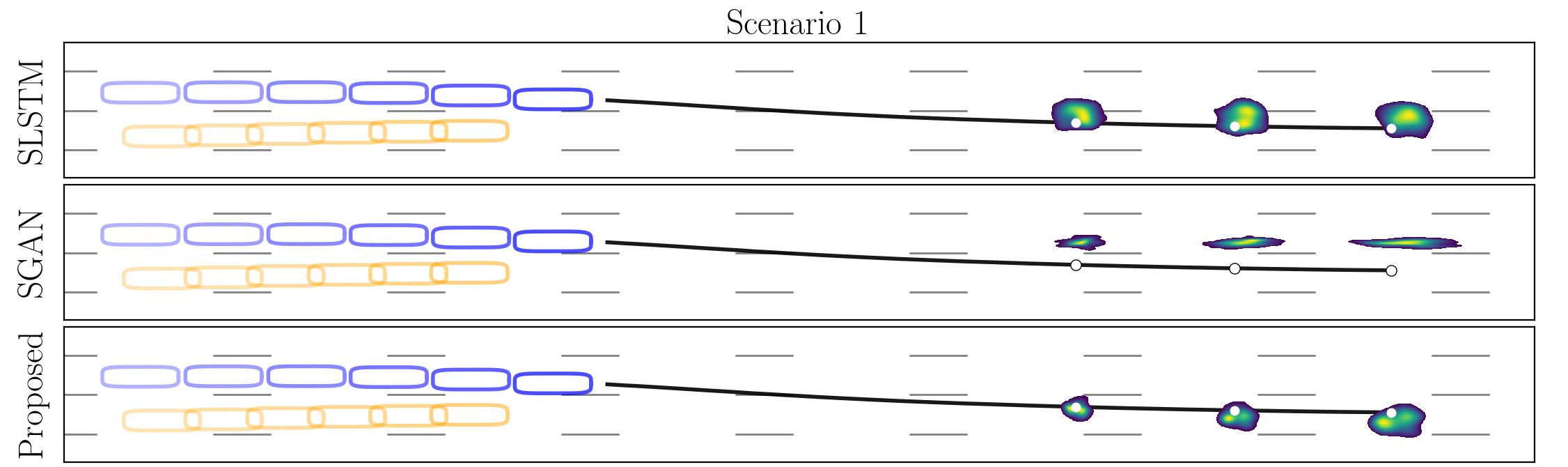}
\includegraphics[width=0.95\textwidth]{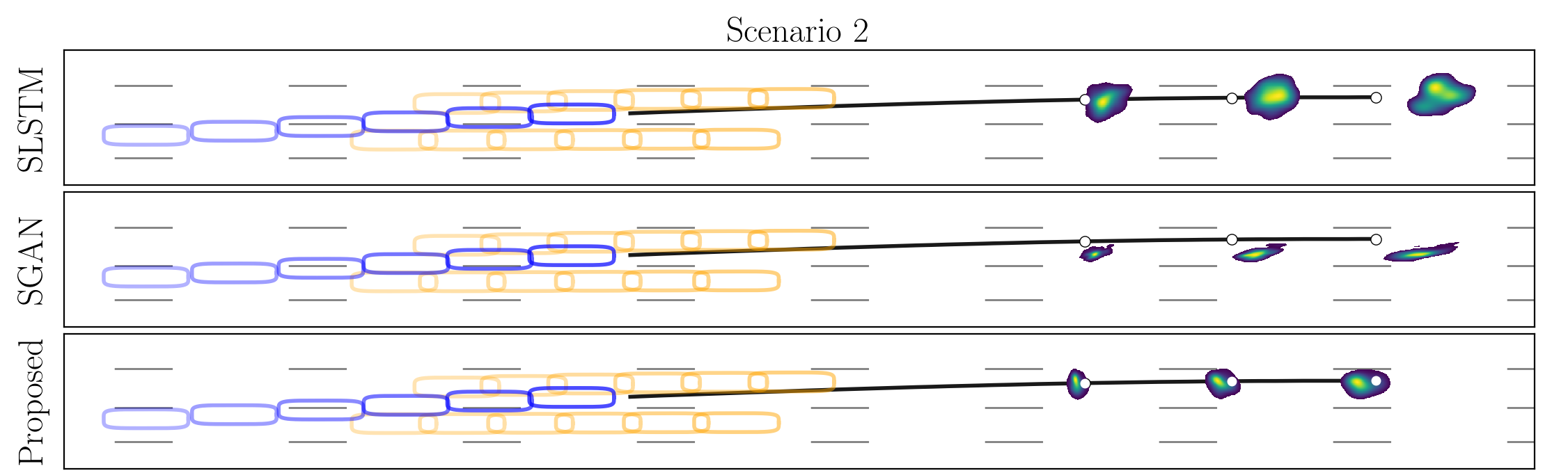}
\includegraphics[width=0.95\textwidth]{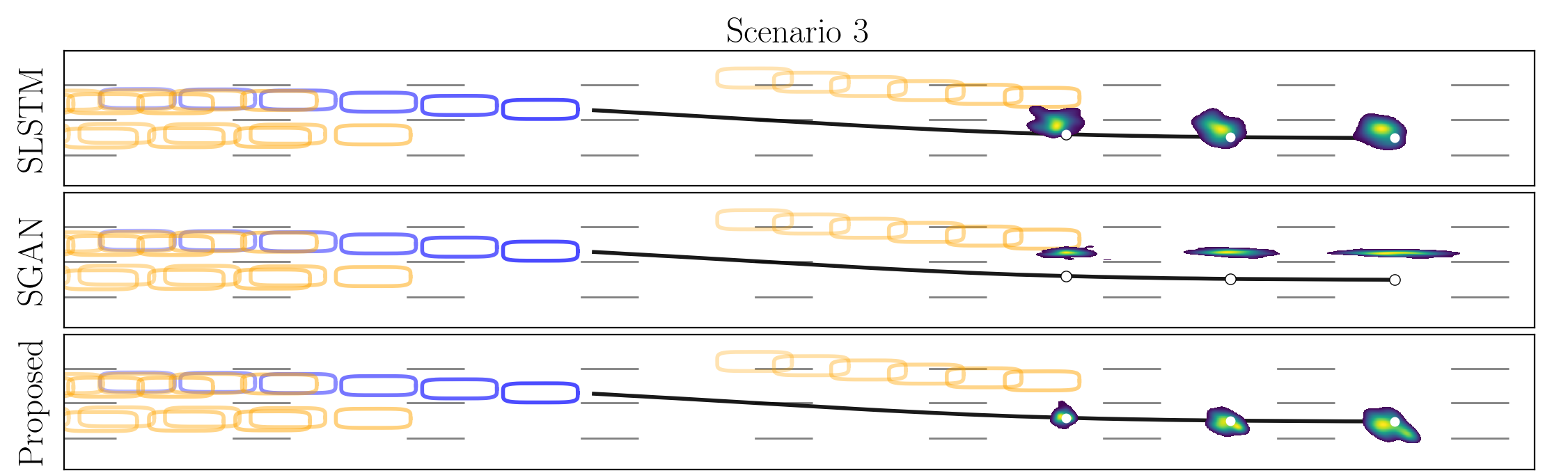}
\centering
\caption{
Examples of bird's-eye view predictions for dynamic scenarios from highD.
Each method observes \SI{3}{\second} of each road user's trajectory (observations shown at each half second) before predicting the next \SI{5}{\second} into the future.
The actual future trajectory is shown in black with white circles marking the driver's position at \SI{3}{\second}, \SI{4}{\second}, and \SI{5}{\second}.
Each method's predictions are shown for these timesteps with likelihood given by the viridis color scale.} \label{fig:predictions}
\end{figure*}

Here we evaluate predictions made with complete observations of all vehicles, as if they were seen from a bird's-eye view. The performance of each method is shown in Table~\ref{tab:traditional_test} \textit{(top)}.
The results show a trade-off between minimizing the quantile error and the expectation based errors.
Social LSTM achieves low quantile error, it does so at the cost of higher average distances.
On the other hand, MATF generates highly accurate but nearly deterministic predictions. The small difference between its quantile distance and average distance errors indicate that the predictions tend to deviate little from the mean prediction.
Social GAN achieves more balance than the previous two methods in minimizing the different types of errors.
Predictions made by the less deterministic methods are shown in Figure~\ref{fig:predictions}.
Despite the proposed method's limited treatment of interactions, it performs competitively with the other methods, outperforming them for highD scenarios.
Considering car-following interaction aids in predicting changes in speed when other vehicles merge into the same lane.
Comparison to the ablated version also shows that considering interactions improves the predictions across all metrics.

\subsection{Driver View Predictions}
Realistic driving conditions create occlusions that prevent a clear view of other drivers.
Autonomous vehicles are additionally subject to limited sensor range.
In this section we simulate these conditions for each vehicle as if it were the autonomous vehicle.
The simulated AV only observes vehicles within \SI{50}{\meter} of its position along the longitudinal axis, that are not occluded by other vehicles.
Occlusions are generated with a simplified model of detections.
This treats vehicles as spherical obstacles with a radius of \SI{2}{\meter}, and a vehicle is considered occluded if the line from its position to the simulated AV's position collides with any obstacles.
We also ensure each observed vehicle is observed for at least one second in total during the observation window.
Since the deep learning baselines assume full observations, we fill the missing values assuming constant velocity.
The proposed method also assumes the longitudinal positions of surrounding vehicles are given, for which we smooth according to the model dynamics~\eqref{eqn:dynamics} with zero control input.
Table~\ref{tab:traditional_test} \textit{(bottom)} shows that the more challenging nature of the partially observed case leads to a drop in performance across all methods.
The proposed method degrades more gracefully than the baselines, and closes the gaps in performance on NGSIM.








\section{Conclusion} \label{sec:conclusion}

We propose a novel kinematic model to describe both car-following and lane changing behavior.
This provides a means of obtaining interpretable trajectory predictions when the leading vehicle and desired lane are known.
Through Bayesian model averaging, we extend the model to predict trajectories for general highway scenarios in which the designation of leader and follower vehicle may be more ambiguous, and the desired lane is not known.
Experiments on the NGSIM and highD datasets demonstrate that the method is competitive and can outperform state-of-the-art prediction methods.
These findings are shown to hold across varied sensing conditions including both perfect sensors and realistic sensors subject to occlusions.
A benefit of the proposed model's interpretable nature also lies in identifying its weaknesses.
The kinematic model does not impose constraints, which is less appropriate for describing vehicles' constrained lateral motion at low velocities.
A possible remedy is to transition to a different model of motion at low velocities similar to~\cite{houenou2013vehicle}.
The proposed model also assumes that longitudinal and lateral motion are approximately decoupled, which is unrealistic for aggressive maneuvers that are limited across both by friction.
Incorporating these more realistic vehicle dynamics and examining other forms of interaction between drivers provide avenues for future research.









\bibliographystyle{IEEEtran}
\bibliography{IEEEabrv,root}

\end{document}